\documentclass[letterpaper, 10 pt, conference]{ieeeconf}

\IEEEoverridecommandlockouts
\newif\ifreview
\reviewfalse

\usepackage{amsmath,amssymb}
\usepackage{bm}
\usepackage{booktabs}
\usepackage{graphicx}
\usepackage{capt-of}
\usepackage{url}
\usepackage{xcolor}
\newcommand{\gvr}{\hspace{\tabcolsep}{\color{black!18}\vrule width 0.3pt}\hspace{\tabcolsep}}%
\usepackage[ruled,vlined,linesnumbered,algo2e]{algorithm2e}
\DontPrintSemicolon
\SetKwComment{Comment}{\(\triangleright\) }{}
\SetAlgoSkip{}
\makeatletter
\renewcommand{\@algoskip}{}%
\g@addto@macro{\@algocf@finish}{\par\nointerlineskip\vspace{-5pt}}%
\makeatother
\usepackage{tikz}
\usetikzlibrary{arrows.meta,positioning,calc,fit,shapes.geometric}

\newtheorem{theorem}{Theorem}

\newtheorem{lemma}{Lemma}

\newtheorem{definition}{Definition}

\newcommand{\Real}{\mathbb{R}}
\newcommand{\Rmin}{\overline{\Real}_{\ge 0}}
\newcommand{\infval}{\infty}

\newcommand{\oneK}{0}
\newcommand{\otime}{\otimes}
\newcommand{\oplusop}{\oplus}
\newcommand{\ports}{B}
\newcommand{\Eqref}[1]{Eq.~\eqref{#1}}

\newcommand{\RepairClosure}{\texttt{RepairClosure}}
\newcommand{\Map}[1]{\ensuremath{M_{#1}}}

\makeatletter
\renewcommand{\fnum@table}{Table~\thetable}
\long\def\@makecaption#1#2{%
  \ifx\@captype\@IEEEtablestring
    \setbox\@tempboxa\hbox{\footnotesize #1.~~ #2}%
    \ifdim \wd\@tempboxa >\hsize
      \setbox\@tempboxa\hbox{\footnotesize #1.~~ }%
      \parbox[t]{\hsize}{\footnotesize \noindent\unhbox\@tempboxa#2}%
    \else
      \hbox to\hsize{\footnotesize\box\@tempboxa\hfil}%
    \fi
    \@IEEEtablecaptionsepspace
  \else
    \@IEEEfigurecaptionsepspace
    \setbox\@tempboxa\hbox{\footnotesize #1.~~ #2}%
    \ifdim \wd\@tempboxa >\hsize
      \setbox\@tempboxa\hbox{\footnotesize #1.~~ }%
      \parbox[t]{\hsize}{\footnotesize \noindent\unhbox\@tempboxa#2}%
    \else
      \ifcenterfigcaptions \hbox to\hsize{\footnotesize\hfil\box\@tempboxa\hfil}%
      \else \hbox to\hsize{\footnotesize\box\@tempboxa\hfil}%
      \fi
    \fi
  \fi
}
\makeatother

\makeatletter
\renewcommand\section{\@startsection{section}{1}{\z@}{1.0ex plus 0.25ex minus 0.4ex}%
{0.50ex plus 0.12ex minus 0.10ex}{\normalfont\normalsize\centering\scshape}}%
\renewcommand\subsection{\@startsection{subsection}{2}{\z@}{0.85ex plus 0.20ex minus 0.30ex}%
{0.40ex plus 0.10ex minus 0.10ex}{\normalfont\normalsize\itshape}}%
\makeatother

\AddToHook{env/lemma/begin}{\setlength{\topsep}{4pt plus 1pt minus 2pt}\setlength{\partopsep}{0pt}}
\AddToHook{env/theorem/begin}{\setlength{\topsep}{4pt plus 1pt minus 2pt}\setlength{\partopsep}{0pt}}
\AddToHook{env/corollary/begin}{\setlength{\topsep}{4pt plus 1pt minus 2pt}\setlength{\partopsep}{0pt}}
\AddToHook{env/definition/begin}{\setlength{\topsep}{4pt plus 1pt minus 2pt}\setlength{\partopsep}{0pt}}
\AtBeginDocument{%
  \abovedisplayskip 1ex plus 2pt minus 1pt\relax
  \belowdisplayskip \abovedisplayskip
  \abovedisplayshortskip 0pt plus 2pt\relax
  \belowdisplayshortskip 1ex plus 2pt minus 1pt\relax
  \jot=2pt
  \setlength{\abovecaptionskip}{3pt plus 1pt minus 1pt}%
  \setlength{\textfloatsep}{4pt plus 0pt minus 1pt}%
  \setlength{\floatsep}{8pt plus 0pt minus 1pt}%
  \setlength{\intextsep}{4pt plus 0pt minus 1pt}%
  \setlength{\dbltextfloatsep}{4pt plus 0pt minus 1pt}%
  \setlength{\dblfloatsep}{8pt plus 0pt minus 1pt}%
  \addtolength{\textfloatsep}{0.2em}%
  \addtolength{\floatsep}{0.2em}%
  \addtolength{\intextsep}{0.2em}%
  \addtolength{\dbltextfloatsep}{0.2em}%
  \addtolength{\dblfloatsep}{0.2em}%
  \setlength{\csname @fptop\endcsname}{0pt}%
  \setlength{\csname @fpsep\endcsname}{\dimexpr 8pt + 0.2em\relax plus 1fil}%
  \setlength{\csname @fpbot\endcsname}{0pt}%
  \setlength{\csname @dblfptop\endcsname}{0pt}%
  \setlength{\csname @dblfpsep\endcsname}{\dimexpr 8pt + 0.2em\relax plus 1fil}%
  \setlength{\csname @dblfpbot\endcsname}{0pt}%
  \typeout{HTRIP-floatsep textfloatsep=\the\textfloatsep}%
}

\title{\LARGE \bf H-TRIP: Exact Many-to-Many Distance Index for \\ Multi-Robot Exploration\thanks{\ifreview The source code is available at \protect\url{https://github.com/ozaslan/htrip}.\else This work has been supported by Ankara Yildirim Beyazit University Scientific Research Projects Coordination Unit under Multidisciplinary Research Project No.~3225. The source code is available at \protect\url{https://github.com/ozaslan/htrip}.\fi}}

\ifreview
\author{Anonymous Authors}
\else
\author{
\authorblockN{Elif Alt{\i}nay \"{O}zaslan}
\authorblockA{Department of Mathematics\\
Kastamonu University, Kastamonu, T\"{u}rkiye\\
{\tt eaozaslan@kastamonu.edu.tr}}
\and
\authorblockN{Tolga \"{O}zaslan}
\authorblockA{Department of Computer Engineering\\
Ankara Y{\i}ld{\i}r{\i}m Beyaz{\i}t University, Ankara, T\"{u}rkiye\\
{\tt tozaslan@aybu.edu.tr}}
}
\fi

\IEEEaftertitletext{\vspace{-1\baselineskip}}

\begin{document}

\maketitle
\thispagestyle{empty}
\pagestyle{empty}

\begin{abstract}
    Frontier-based exploration systems repeatedly evaluate geodesic travel costs between robot poses and candidate viewpoints, and tour-based planners consume the full pairwise distance matrix built from them.
    Recomputing these matrices with one graph search per source scales with the explored map size, whereas distance caches that refresh only locally affected or newly created frontiers can retain stale costs between surviving pairs when newly discovered free space creates a global shortcut.
    We present H-TRIP (Hierarchical Tropical Rank-one Incremental Propagation), an exact distance index grounded in Kleene algebra over the min-plus semiring for monotonically growing occupancy grids.
    H-TRIP partitions the known map into a spatial hierarchy with fixed boundary ports.
    Each unknown-to-free revelation is decomposed into efficient rank-one min-plus edge insertions at the affected leaf, and only decreased port-to-port summaries are propagated along its ancestor chain.
    A shared hierarchical lift evaluates the dense many-to-many matrix of shortest distances for all active query points on the occupancy map.
    We prove that sequential insertions preserve exact shortest-path distances on the grid.
    Benchmark experiments confirm exact agreement with reference searches on fully built maps and show that dense many-to-many queries at a few dozen viewpoints are up to two orders of magnitude faster than sequential \(k\times\)BFS.
    On exploration runs with sixteen robots, in-loop queries on the growing occupancy graph are \(2.21\times\)--\(8.59\times\) faster than co-scheduled \(k\times\)BFS on maps with side \(256\), \(512\), and \(1024\) cells.
\end{abstract}

\noindent\textit{Index Terms}---Multi-robot exploration, APSP, distance indexes, min-plus algebra, occupancy grids.

\suppressfloats[t]
\begin{figure}[t]
    \setlength{\abovecaptionskip}{4pt}
    {\centering
        \begin{minipage}[t]{0.32\columnwidth}
            \centering
            \includegraphics[width=\linewidth]{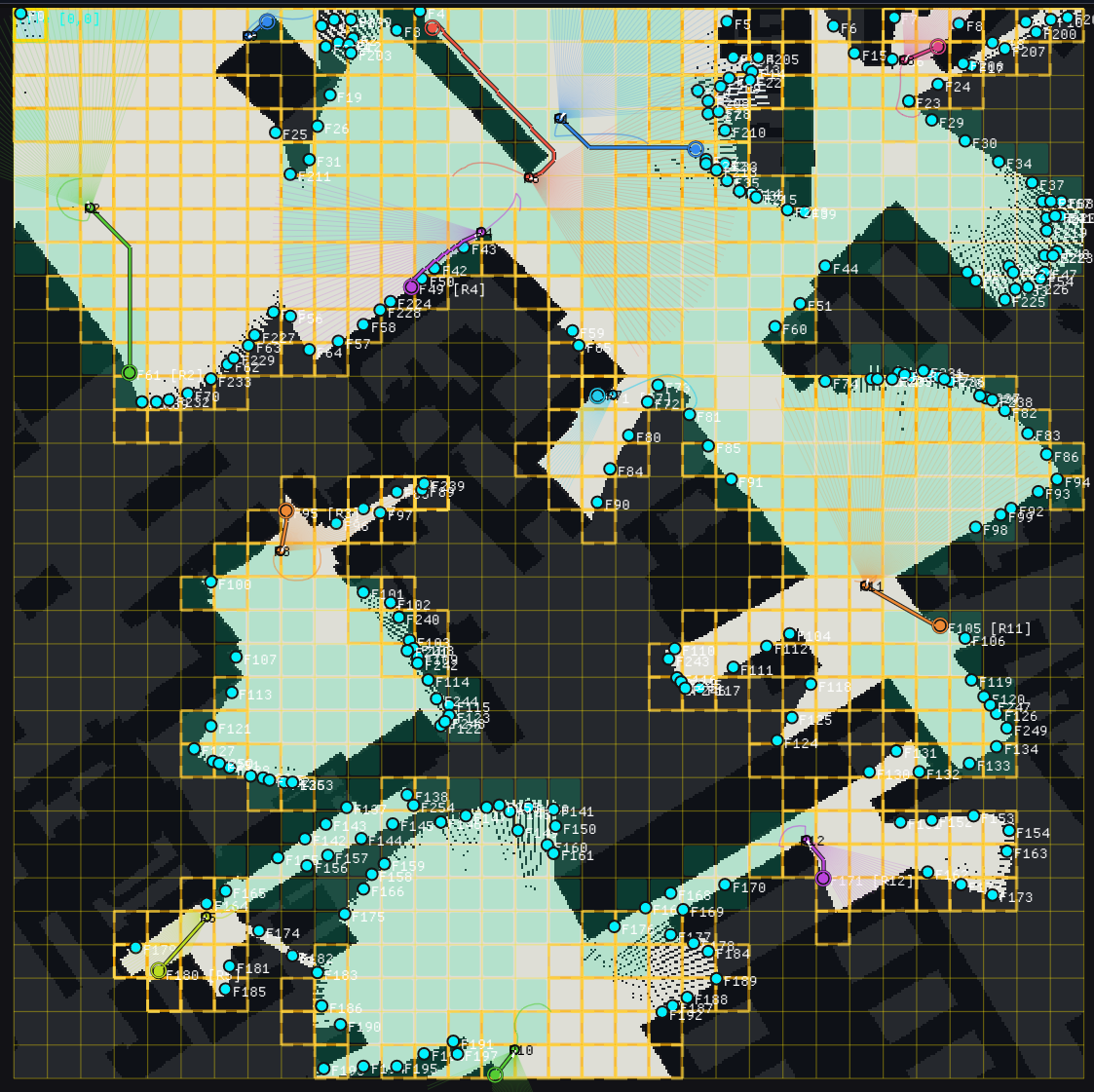}
        \end{minipage}\hfill
        \begin{minipage}[t]{0.32\columnwidth}
            \centering
            \includegraphics[width=\linewidth]{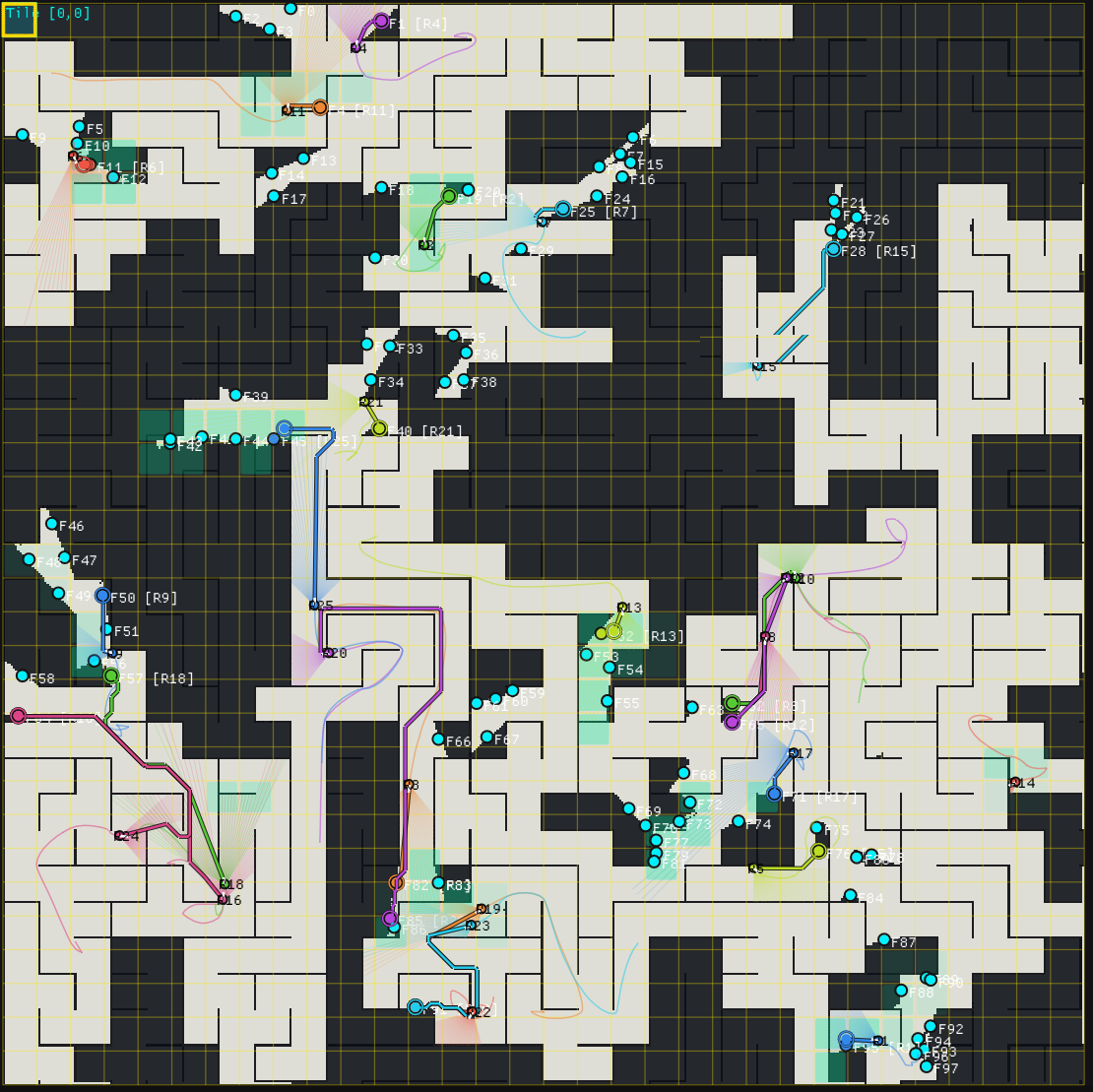}
        \end{minipage}\hfill
        \begin{minipage}[t]{0.32\columnwidth}
            \centering
            \includegraphics[width=\linewidth]{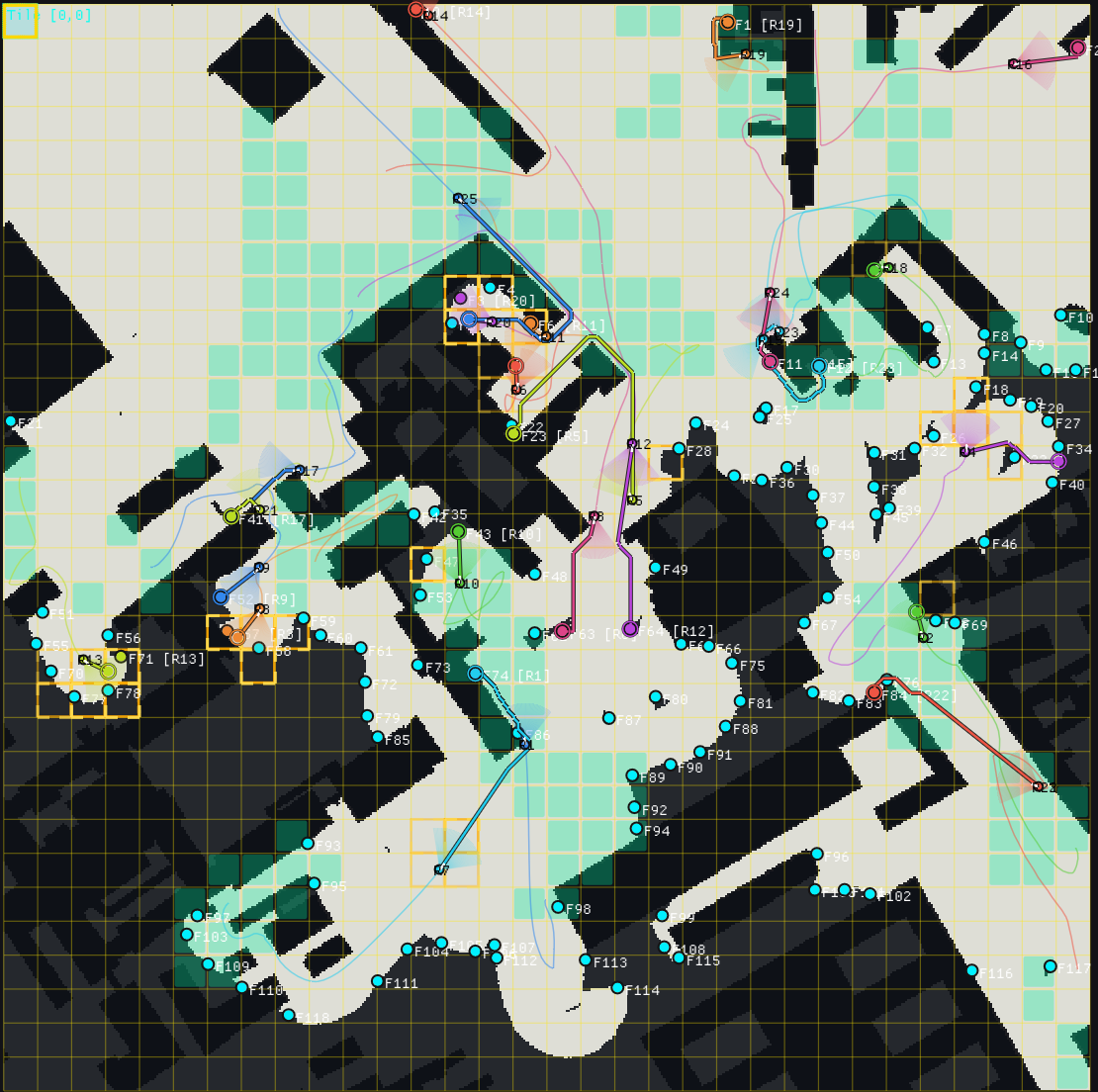}
        \end{minipage}\par}
    \caption{We tested H-TRIP in simulation, where lidar-equipped robot teams of different sizes explored occupancy grids of different side lengths.
        The panels show three representative frames from these runs.
        The yellow lattice is the leaf-tile partition, cyan markers are active frontiers, and colored polylines are assigned open tours.}
    \label{fig:explorer}
\end{figure}

\section{Introduction}
\label{sec:intro}

Autonomous exploration is foundational in search and rescue, inspection, and reconnaissance.
Mobile robot teams iteratively map free space, extract candidate frontier viewpoints, and coordinate assignments using travel cost, information gain, or both~\cite{burgard2005coordinated}.
Tour-based planners repeatedly evaluate pairwise travel costs among candidate goals using geodesic distance as their spatial backbone, which can be combined with planner-specific kinematic or utility terms~\cite{zhou2021fuel,cao2021tare,zhou2023racer}.
These geodesics are shortest paths on the free-space graph \(G_\tau=(V_\tau,E_\tau)\) of an occupancy representation or an exploration roadmap.
Under undirected traversal the geodesic distance matrix is symmetric, so filling \(D\in\Rmin^{k\times k}\) among \(k\) robot poses and candidate viewpoints with the \(k(k-1)/2\) distinct pairwise distances of \(G_\tau\) requires substantial computational effort as maps and candidate sets grow.

As sensors reveal free space, the traversable graph \(G_\tau\) grows monotonically.
Recomputing \(D\) from scratch with one graph search per source scales with both \(k\) and \(|V_\tau|\), and competes with the planning cycle.
Newly revealed free cells can complete a doorway or corridor that shortens the true geodesic between two persisting viewpoints, making a previously cached distance stale.
The challenge is to keep the geodesic distance matrix \(D\) exact without recomputing shortest paths over the full graph after every occupancy-map update.
We therefore ask how a multi-robot system can maintain exact all-pairs geodesic distances among surviving viewpoints under online monotonic map growth while keeping many-to-many queries within a planning-cycle budget.

We present H-TRIP (Hierarchical Tropical Rank-one Incremental Propagation), an exact hierarchical distance index grounded in Kleene algebra over the min-plus semiring~\cite{conway1971regular,rote1990path} that maintains pairwise geodesic distances as occupancy grows and answers many-to-many queries for assignment and tour planners on online occupancy grids.
H-TRIP organizes the workspace into a balanced spatial hierarchy over fixed leaf tiles whose boundary-port sets remain fixed throughout exploration.
Sensor measurements that reveal previously unknown cells as free insert local edges in \(G_\tau\), including newly passable cross-tile seams, and only decreased boundary-distance summaries propagate upward along the affected ancestor chain.

H-TRIP builds this hierarchy so that many-to-many geodesic distances can be evaluated by lifting each of the \(k\) viewpoints all at once onto the boundary ports of their ancestor tiles and resolving pairwise distances as tropical products against stored boundary metrics.
H-TRIP realizes its query advantage by amortizing each hierarchical lift across the entire \(k\times k\) matrix.
An isolated pair lookup pays that lift without sharing it, so batched many-to-many evaluation is the regime that reuses the work.
Hence H-TRIP can serve as a metric backend for multi-robot assignment and tour planners that require geodesic distance.



In simulation, H-TRIP consistently preserved exact geodesic distances while reducing the cost of repeated many-to-many evaluation relative to conventional graph-search baselines.
The advantage was most pronounced when multiple robot poses and frontier viewpoints were queried together, allowing hierarchical computations to be shared across the complete distance matrix.

The main contributions of this paper are:
(i) an exact hierarchical distance index for monotonically growing occupancy grids based on fixed geometric interfaces and min-plus closure;
(ii) a localized insertion-only maintenance procedure based on rank-one min-plus updates, with exactness established by Lemma~1 and Theorem~1; and
(iii) a batched hierarchical query that recovers the complete \(k\times k\) geodesic distance matrix, with exactness established by Theorem~2.
\section{Related Work}
\label{sec:related}

The literature most relevant to H-TRIP spans four areas: exploration planning and travel-cost maintenance, incremental replanning, hierarchical distance indexes, and dynamic shortest-path maintenance.

\vspace{0.2em}
\noindent\textit{Exploration planners and travel-cost maintenance:}
Frontier exploration guides robots toward boundaries between mapped free space and unmapped terrain, balancing travel distance against information gain~\cite{burgard2005coordinated}.
Tour-based planners such as FUEL~\cite{zhou2021fuel}, TARE~\cite{cao2021tare}, and RACER~\cite{zhou2023racer} consume pairwise travel costs over robot poses and candidate goals (frontiers), which makes geodesic path length their spatial backbone.
When a new frontier cluster appears, FUEL~\cite{zhou2021fuel} searches paths from that cluster's candidate viewpoints and reuses cached costs for frontier clusters that did not change, so a pair of robot poses or frontier viewpoints that survive the update can retain a pre-opening length.
TARE~\cite{cao2021tare} localizes traveling salesperson (TSP) work through spatial partitions.
As the map grows, a newly opened doorway or corridor can shorten the true geodesic between surviving viewpoints, which can cause \emph{cached} surviving-pair geodesics to become stale.
H-TRIP does not replace those assignment or tour layers.
It supplies the geodesic distance matrix they consume and keeps that matrix \emph{exact} as the map grows, so a surviving-pair shortcut is reflected without a global graph search at every update.

\noindent\textit{Incremental single-source replanning:}
In partially known environments, D\(^*\) Lite~\cite{koenig2002dstar} and LPA\(^*\)~\cite{koenig2004lpa} incrementally repair a single-source or goal-rooted search tree as edge costs change.
A straightforward use of these single-source methods for \(k\) viewpoints requires maintaining or repairing multiple search states, typically one per source.
H-TRIP is not a path replanner and does not maintain a previously found tree.
It maintains one hierarchical metric shared by all queried viewpoints and fills the dense \(k\times k\) distance matrix by batched coordinate lifts and tropical products, leaving path construction to the tour layer that consumes those distances.

\noindent\textit{Hierarchical distance indexes:}
Spatial hierarchies compress distance computation by storing summaries on region boundaries.
HPA\(^*\)~\cite{botea2004near} partitions a grid into clusters, identifies data-dependent border entrances on traversable boundaries, and caches intra-cluster entrance distances for search on an abstract graph, which returns a \emph{near-optimal} path rather than an exact stored distance.
HiTi~\cite{jung2002hiti} is closer to H-TRIP because it recursively partitions road graphs, stores precomputed minimum-cost paths among boundary nodes, and supports optimal shortest-path queries, with updates confined to ancestor path-view sets.
However, in the online occupancy setting H-TRIP fixes ports geometrically before exploration, applies rank-one min-plus insertions on leaf and interface closures when cells open, propagates only decreased ancestor metrics, and evaluates the dense \(k\times k\) matrix with one batched hierarchical lift shared by all viewpoints.

On road networks and static graphs, Contraction Hierarchies~\cite{geisberger2012contraction} add vertex shortcuts, whereas Customizable Route Planning (CRP)~\cite{delling2017customizable} and Customizable Contraction Hierarchies~\cite{dibbelt2016customizable} customize metric overlays on separator cliques or nested-dissection orders.
Dynamic shortest-path indexes also admit streaming maintenance under weight updates~\cite{ouyang2020efficient}.
Those methods assume a known topology whose weights change.
Exploration instead changes topology: newly revealed free cells insert vertices and edges that were absent from the occupancy graph.
H-TRIP still keeps the tile partition and port universe geometrically fixed, independent of current passability, so an opening does not resize overlays or churn entrances.

Dense distance graphs on recursive \(r\)-divisions support exact distances among all boundary vertices of a region inside static planar algorithms~\cite{fakcharoenphol2006planar}.
On a static preprocessed road network whose topology is already known, many-to-many tables share hierarchical search work across a source set and a target set, so filling an \(|S|\times|T|\) table is cheaper than a separate query per pair~\cite{knopp2007many,delling2011rphast}.
H-TRIP stores tile-local port summaries and lifts only the \(k\) active viewpoints, such that \(k\ll |V_\tau|\), and the planner consumes a \(k\times k\) geodesic distance matrix produced by one batched hierarchical lift.

\noindent\textit{Dynamic APSP and closed-semiring algebra:}
Demetrescu and Italiano~\cite{demetrescu2004dynamic} maintain fully dynamic exact all-pairs shortest paths on general directed graphs with non-negative edge weights, covering both insertions and deletions.
H-TRIP applies the same rank-one decrease~\cite{conway1971regular,lehmann1977algebraic,rote1990path} to leaf and interface closures along the affected ancestor chain.
Its exactness theorem is insertion-only occupancy growth rather than fully dynamic APSP, and its query is a batched \(k\times k\) hierarchical lift rather than an \(O(1)\) pair lookup.

Ramalingam and Reps~\cite{ramalingam1996incremental} repair incremental shortest paths on the affected region, and G{\'o}rkiewicz and Karczmarz~\cite{gorkiewicz2025incremental} study approximate incremental APSP on sparse directed graphs.
In its insertion-only occupancy-grid setting, H-TRIP remains exact by confining dense min-plus updates to the affected ancestor chain, rather than an affected-region search or an approximate APSP.

Collectively, prior work provides incremental frontier-cost caching, single-source replanning, hierarchical shortest-path representations, and general dynamic distance maintenance.
H-TRIP combines (i)~fixed geometric interfaces, (ii)~insertion-only exact min-plus updates, and (iii)~localized hierarchical propagation into one index, and \emph{evaluates} the \(k\times k\) geodesic distance matrix from that index, with \(k\ll |V_\tau|\) for medium-to-large maps.

\section{Problem Formulation}
\label{sec:problem}

\subsection{Occupancy Grid and Monotonic Map Growth}
Let \(\mathcal{W}\subset\mathbb{R}^2\) be a planar workspace discretized into a uniform occupancy grid \(M\).
Each cell is labeled \(\texttt{unknown}\), \(\texttt{free}\), or \(\texttt{obstacle}\).
Confirmed \(\texttt{free}\) cells induce an undirected graph \(G_\tau=(V_\tau,E_\tau)\), where \(V_\tau\) is the set of currently free cells and \(\tau\) is a discrete time index of successive occupancy graphs.
By default \(E_\tau\) joins \(4\)-adjacent free cells with unit weight, so geodesics are integers.
Non-unit nonnegative cell costs, when present, become incident edge weights, and the reference search is Dijkstra's algorithm rather than BFS.
Let \(d_{G_\tau}(u,v)\) denote shortest-path length on \(G_\tau\), and \(\infty\) if \(v\) is unreachable from \(u\).
Unknown cells retain geometric slots with infinite incident weight until they are revealed free, and obstacle cells remain blocked.

An \emph{opening} is a set \(\Delta V\) of cells that become free, together with the finite incident edges \(\Delta E\) that those cells insert.
Hence \(V_{\tau+1}=V_\tau\cup\Delta V\) and \(E_{\tau+1}=E_\tau\cup\Delta E\), and distances can only decrease or stay the same.
We consider exploration of a static workspace in which sensing monotonically reveals previously unknown space.

\subsection{Robots, Frontiers, and the Geodesic Distance Matrix}
\label{sec:planning_matrix}
A frontier cell is \(\texttt{free}\) and adjacent to \(\texttt{unknown}\).
Connected \(4\)-adjacent frontier cells form clusters \(\mathcal{F}_\tau=\{F_1,\dots,F_K\}\), each represented by a medoid cell \(m(F_i)\)~\cite{zhou2021fuel,cao2021tare}.
Each robot contributes a planar position \(r_i\in\mathcal{W}\), which Section~\ref{sec:system} snaps onto a free cell \(\tilde{r}_i\in V_\tau\), so every finite entry of \(D_\tau\) is a grid geodesic among cells of \(G_\tau\).

A planning cycle assembles the viewpoint list
\begin{equation}
    \label{eq:viewpoint_list}
    P_\tau = \bigl(\tilde{r}_1,\dots,\tilde{r}_N,\; m(F_1),\dots,m(F_K)\bigr),
\end{equation}
with \(k=|P_\tau|=N+K\), which may change between cycles as clusters appear, vanish, or move.
The geodesic distance matrix \(D_\tau\in\Rmin^{k\times k}\) is
\begin{equation}
    \label{eq:D_matrix}
    (D_\tau)_{ij} = d_{G_\tau}(P_{\tau,i}, P_{\tau,j}),
\end{equation}
with blocks
\begin{equation}
    \label{eq:D_blocks}
    D_\tau =
    \begin{bmatrix}
        D_{\tau,RR} & D_{\tau,RF} \\
        D_{\tau,FR} & D_{\tau,FF}
    \end{bmatrix}
\end{equation}
for robot--robot, robot--frontier, and frontier--frontier distances.
Traversal is undirected, so \(D_\tau\) is symmetric.
Without a distance index, filling \(D_\tau\) requires \(k\) single-source searches on \(G_\tau\) (BFS if unweighted, Dijkstra if weighted), or \(k(k-1)/2\) independent point-to-point \(\mathrm{A}^*\) searches.

In a typical multi-robot exploration system, assignment looks up robot--frontier costs in \(D_{\tau,RF}\), and an open tour looks up frontier--frontier costs in \(D_{\tau,FF}\).
Together those stages consume the \(D_{\tau,RF}\) and \(D_{\tau,FF}\) blocks of the geodesic matrix.
The same query returns the full symmetric \(k\times k\) matrix, so \(D_{\tau,RR}\) is filled with the robot--robot geodesics.
If an entry of \(D_{\tau,RF}\) or \(D_{\tau,FF}\) is stale, assignment or visit order can change after a doorway opens, which is why planner-facing entries of \(D_\tau\) must remain exact on the occupancy graph after each absorbed \emph{opening}.

\subsection{The Surviving-Pair Shortcut Problem}
After an \emph{opening} that produces \(G_{\tau+1}\), the true geodesic between two query cells that remain in the planning list can become strictly shorter than the length that used to be correct on \(G_\tau\).
A viewpoint persists in the cache only when its grid coordinate is unchanged.
An incremental \emph{new-cluster-only cache} recomputes pairwise distances only for newly appeared clusters and, for speed, leaves the remaining entries unchanged, which can leave those lengths stale after an opening~\cite{zhou2021fuel}.

\begin{definition}[Surviving-Pair Shortcut]
    \label{def:shortcut}
    A surviving-pair shortcut occurs when two viewpoints occupy the same cells in \(P_\tau\) and in \(P_{\tau+1}\) and, after the edges \(\Delta E=E_{\tau+1}\setminus E_\tau\) of the opening, \(G_{\tau+1}\) contains a path between them that is strictly shorter than the distance stored for that pair in \(D_\tau\).
\end{definition}

Such a caching mechanism is the failure mode under study, because a retained entry of its matrix \(\widehat{D}_{\tau+1}\) can satisfy \((\widehat{D}_{\tau+1})_{ij}=(D_\tau)_{ij}>(D_{\tau+1})_{ij}\) even when both cells lie far from the newly free vertices \(\Delta V\).

\subsection{Problem Statement}
A distance index is a map \(\mathcal{I}:V_\tau^{\kappa}\to\Rmin^{\kappa\times\kappa}\) whose input is an ordered list of \(\kappa\) cells of \(G_\tau\) and whose output is the geodesic distance matrix on that list.
Given a monotonically expanding occupancy graph \(G_\tau\to G_{\tau+1}\) and a viewpoint list \(P_{\tau+1}\in V_{\tau+1}^{k}\), we ask for an index \(\mathcal{I}\) that meets one mathematical contract and two computational objectives:
(1)~exactness: after newly free cells have been absorbed, \(\mathcal{I}(P_{\tau+1})\) equals \(D_{\tau+1}\) entrywise on that graph;
(2)~update locality: the work that absorbs an opening is confined to the affected leaf tiles and the union of their ancestor chains;
(3)~many-to-many output: a single query returns \(D_{\tau+1}\).

\begin{figure}[!t]
\centering
\resizebox{\columnwidth}{!}{%
\begin{tikzpicture}[
  x=1mm, y=1mm,
  font=\scriptsize,
  line width=0.35pt,
  cell/.style={draw=black!25, fill=black!3, line width=0.20pt},
  obs/.style={draw=black!55, fill=black!45, line width=0.22pt},
  extcell/.style={draw=portblue!80!black, fill=portblue, line width=0.35pt},
  seamcell/.style={draw=portblue, fill=portblue!22, line width=0.35pt},
  sarr/.style={-{Latex[length=1.9pt, width=1.4pt]}, line width=0.58pt, draw=seamgold},
  sarrbi/.style={{Latex[length=1.9pt, width=1.4pt]}-{Latex[length=1.9pt, width=1.4pt]}, line width=0.58pt, draw=seamgold},
  card/.style={draw=black!18, fill=black!1, rounded corners=1.0mm, line width=0.35pt}
]
\definecolor{portblue}{RGB}{24,102,175}
\definecolor{seamgold}{RGB}{230,150,10}

\draw[card] (0, 5.2) rectangle (32, 37.0);
\node[font=\footnotesize\bfseries, text=black!85] at (16, 34.0) {(\textbf{a}) Base tile \(T\) (\(b=5\))};

\begin{scope}[shift={(9.25, 14.2)}]
  \def\csa{2.7}
  \foreach \i in {0,...,4} {
    \foreach \j in {0,...,4} {
      \draw[cell] (\i*\csa, \j*\csa) rectangle ++(\csa, \csa);
    }
  }
  \draw[obs] (2*\csa, 2*\csa) rectangle ++(\csa, \csa);
  \draw[obs] (1*\csa, 3*\csa) rectangle ++(\csa, \csa);

  \foreach \i in {0,...,4} {
    \draw[extcell] (\i*\csa, 0) rectangle ++(\csa, \csa);
    \draw[extcell] (\i*\csa, 4*\csa) rectangle ++(\csa, \csa);
    \draw[extcell] (0, \i*\csa) rectangle ++(\csa, \csa);
    \draw[extcell] (4*\csa, \i*\csa) rectangle ++(\csa, \csa);
  }
  \draw[line width=0.55pt, draw=black!80] (0,0) rectangle (5*\csa, 5*\csa);
  \node[font=\footnotesize\bfseries, text=black!70, anchor=south west] at (-0.3, 5*\csa + 0.3) {\(T\)};
\end{scope}

\draw[-{Latex[length=1.4pt, width=1.1pt]}, draw=portblue!90!black, line width=0.38pt]
  (5.2, 20.95) -- (8.95, 20.95);
\node[font=\scriptsize\bfseries, text=portblue!90!black, anchor=east] at (5.0, 20.95) {\(B_T\)};

\node[font=\scriptsize, text=black!75] at (16, 10.3) {closure \(C_T\) on \(b^2\) cells};
\node[font=\scriptsize, text=black!75] at (16, 7.1) {summary \(S_T\) on \(B_T\)};

\draw[card] (34.0, 5.2) rectangle (88, 37.0);
\node[font=\footnotesize\bfseries, text=black!85] at (61.0, 34.0) {(\textbf{b}) Super-tile \(U\) (\(g=2\))};

\newcommand{\drawuchild}[7]{%
  \begin{scope}[shift={(#1,#2)}]
    \def\cs{1.85}
    \fill[black!3] (\cs, \cs) rectangle (4*\cs, 4*\cs);
    \draw[draw=black!30, line width=0.25pt] (\cs, \cs) rectangle (4*\cs, 4*\cs);
    \node[font=\scriptsize, text=black!60] at (2.5*\cs, 2.5*\cs) {#3};

    \foreach \kk in {0,...,4} {
      \if s#4 \draw[seamcell] (\kk*\cs, 4*\cs) rectangle ++(\cs, \cs); \fi
      \if s#5 \draw[seamcell] (4*\cs, \kk*\cs) rectangle ++(\cs, \cs); \fi
      \if s#6 \draw[seamcell] (\kk*\cs, 0) rectangle ++(\cs, \cs); \fi
      \if s#7 \draw[seamcell] (0, \kk*\cs) rectangle ++(\cs, \cs); \fi
    }
    \foreach \kk in {0,...,4} {
      \if e#4 \draw[extcell] (\kk*\cs, 4*\cs) rectangle ++(\cs, \cs); \fi
      \if e#5 \draw[extcell] (4*\cs, \kk*\cs) rectangle ++(\cs, \cs); \fi
      \if e#6 \draw[extcell] (\kk*\cs, 0) rectangle ++(\cs, \cs); \fi
      \if e#7 \draw[extcell] (0, \kk*\cs) rectangle ++(\cs, \cs); \fi
    }
    \draw[line width=0.45pt, draw=black!75] (0,0) rectangle (5*\cs, 5*\cs);
  \end{scope}
}

\def\oxA{49.8}
\def\oxB{62.8}
\def\oyBot{7.9}
\def\oyTop{20.9}
\def\cs{1.85}
\def\cw{9.25} 

\draw[draw=portblue!70!black, fill=portblue!2, dashed, rounded corners=0.8mm, line width=0.45pt]
  ({\oxA-1.2}, {\oyBot-1.2}) rectangle ({\oxB+\cw+1.2}, {\oyTop+\cw+1.2});

\drawuchild{\oxA}{\oyTop}{\(S_{T_1}\)}{e}{s}{s}{e}
\node[font=\footnotesize\bfseries, text=black!70, anchor=east]
  at ({\oxA-1.8}, {\oyTop+0.5*\cw}) {\(T_1\)};

\drawuchild{\oxB}{\oyTop}{\(S_{T_2}\)}{e}{e}{s}{s}
\node[font=\footnotesize\bfseries, text=black!70, anchor=west]
  at ({\oxB+\cw+1.8}, {\oyTop+0.5*\cw}) {\(T_2\)};

\drawuchild{\oxA}{\oyBot}{\(S_{T_3}\)}{s}{s}{e}{e}
\node[font=\footnotesize\bfseries, text=black!70, anchor=east]
  at ({\oxA-1.8}, {\oyBot+0.5*\cw}) {\(T_3\)};

\drawuchild{\oxB}{\oyBot}{\(S_{T_4}\)}{s}{e}{e}{s}
\node[font=\footnotesize\bfseries, text=black!70, anchor=west]
  at ({\oxB+\cw+1.8}, {\oyBot+0.5*\cw}) {\(T_4\)};

\def\sgxL{\oxA+\cw+0.38}
\def\sgxR{\oxB-0.38}
\draw[sarrbi] (\sgxL, {\oyTop+1.5*\cs}) -- (\sgxR, {\oyTop+1.5*\cs});
\draw[sarr]   (\sgxL, {\oyTop+2.5*\cs}) -- (\sgxR, {\oyTop+2.5*\cs});
\draw[sarrbi] (\sgxL, {\oyTop+3.5*\cs}) -- (\sgxR, {\oyTop+3.5*\cs});
\draw[sarr]   (\sgxL, {\oyBot+1.5*\cs}) -- (\sgxR, {\oyBot+1.5*\cs});
\draw[sarrbi] (\sgxL, {\oyBot+2.5*\cs}) -- (\sgxR, {\oyBot+2.5*\cs});
\draw[sarr]   (\sgxR, {\oyBot+3.5*\cs}) -- (\sgxL, {\oyBot+3.5*\cs});

\def\sgyB{\oyBot+\cw+0.38}
\def\sgyT{\oyTop-0.38}
\draw[sarrbi] ({\oxA+1.5*\cs}, \sgyT) -- ({\oxA+1.5*\cs}, \sgyB);
\draw[sarr]   ({\oxA+2.5*\cs}, \sgyB) -- ({\oxA+2.5*\cs}, \sgyT);
\draw[sarrbi] ({\oxA+3.5*\cs}, \sgyT) -- ({\oxA+3.5*\cs}, \sgyB);
\draw[sarr]   ({\oxB+1.5*\cs}, \sgyT) -- ({\oxB+1.5*\cs}, \sgyB);
\draw[sarrbi] ({\oxB+2.5*\cs}, \sgyB) -- ({\oxB+2.5*\cs}, \sgyT);
\draw[sarrbi] ({\oxB+3.5*\cs}, \sgyT) -- ({\oxB+3.5*\cs}, \sgyB);

\begin{scope}[shift={(0.0, 1.2)}]
  \draw[extcell] (1.0, -0.2) rectangle ++(2.4, 2.4);
  \node[font=\scriptsize, text=black!85, anchor=west] at (4.0, 1.0)
    {Exposed \(B_T, B_U\)};

  \draw[seamcell] (25.0, -0.2) rectangle ++(2.4, 2.4);
  \node[font=\scriptsize, text=black!85, anchor=west] at (28.0, 1.0)
    {Seams \(V_U\setminus B_U\)};

  \draw[sarrbi] (50.5, 1.0) -- (55.0, 1.0);
  \node[font=\scriptsize, text=black!85, anchor=west] at (56.0, 1.0)
    {Seam edges};

  \draw[obs] (71.5, -0.2) rectangle ++(2.4, 2.4);
  \node[font=\scriptsize, text=black!85, anchor=west] at (74.5, 1.0)
    {Obstacle};
\end{scope}

\end{tikzpicture}}
\caption{Fixed-boundary spatial hierarchy.
    (\textbf{a}) Base tile \(T\) (\(b=5\)) stores intra-tile closure \(C_T\) on all cells and exposes summary \(S_T\) on perimeter \(B_T\).
    (\textbf{b}) Super-tile \(U\) (\(g=2\)) aggregates child boundaries into \(V_U\): interior cells are condensed into summaries \(S_{T_i}\), dark blue perimeter cells form exposed parent ports \(B_U\), and light blue cells are interior seams where gold arrows mark passable cross-child edges.}
\label{fig:hierarchy}
\end{figure}
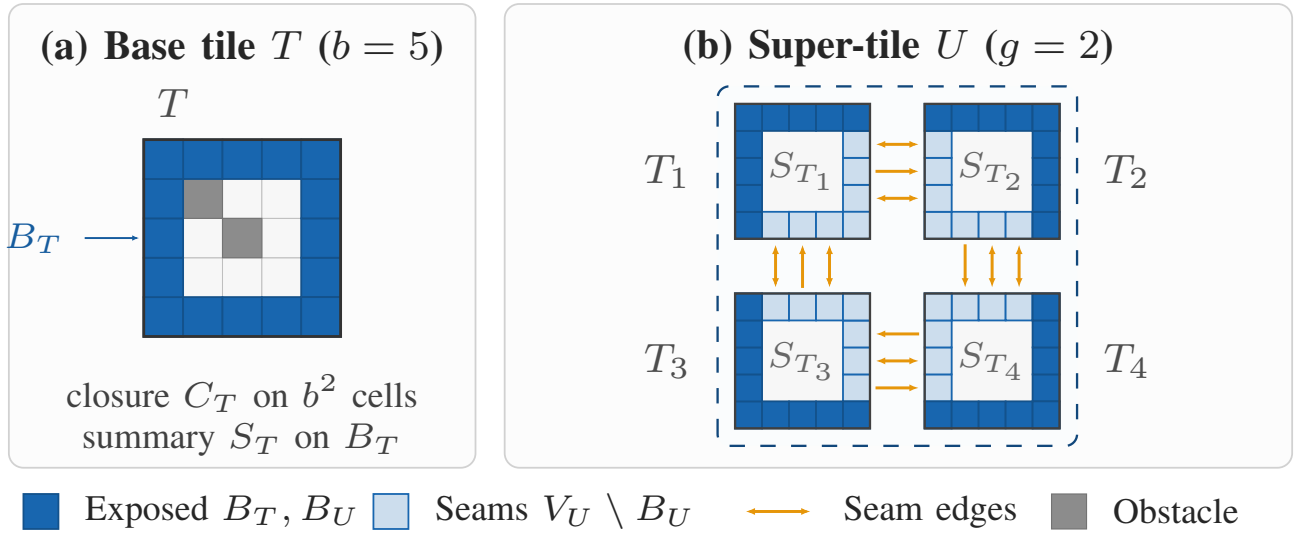
\section{Min-Plus Shortest Paths}
\label{sec:kleene}

Shortest-path computation is expressed over the min-plus semiring \((\Rmin,\min,+,\infval,\oneK)\), where \(a\oplusop b=\min(a,b)\) selects the shorter alternative and \(a\otime b=a+b\) concatenates consecutive path segments~\cite{conway1971regular,lehmann1977algebraic,rote1990path}.
For matrices,
\begin{equation}
  \label{eq:minplus_product}
  (A\otime B)_{ij}=\min_h\bigl(A_{ih}+B_{hj}\bigr).
\end{equation}
If \(A\) is the weighted adjacency matrix of a graph \((V,E)\) with nonnegative edge weights, and \(I_{|V|}\) is the \(|V|\times|V|\) min-plus identity with zeros on the diagonal and \(\infty\) off it, then
\begin{equation}
  \label{eq:kleene}
  A^*=I_{|V|}\oplusop A\oplusop A^{\otime 2}\oplusop\cdots\oplusop A^{\otime(|V|-1)}
\end{equation}
is its all-pairs shortest-path matrix, since a shortest path can be chosen simple and therefore contains at most \(|V|-1\) edges.
The carrier of that algebra is \(\Rmin=\Real_{\ge 0}\cup\{\infty\}\).
H-TRIP applies this algebra locally: tile closures, incremental decreases, hierarchical lifts, and boundary-to-boundary composition are min-plus operations on these matrices.

\section{The H-TRIP Distance Index}
\label{sec:method}
\makeatletter
\newcommand{\ivparlabel}[1]{\protected@edef\@currentlabel{\thesubsection}\label{#1}}
\makeatother

\subsection{Fixed-Boundary Spatial Hierarchy}
The occupancy grid \(M\) is a square \(n\times n\) array of cells, partitioned recursively into square base tiles of side \(b\ge 2\) and super-tiles of tile aggregation degree \(g\ge 2\) whose children form a \(g\times g\) block.
The independent parameters are \(n\), \(b\), and \(g\), from which the depth and the root side follow,
\begin{equation}
  \label{eq:depth}
  H=\lceil\log_g(n/b)\rceil,\qquad b_H=bg^{H}\ge n,
\end{equation}
and every cell of \(G_\tau\) lies inside that root, so \(|V_\tau|\le b_H^2\).

\begin{definition}[Fixed Boundary]
  For a level-\(\ell\) tile \(T\) of side \(b_\ell=bg^{\ell}\), the boundary-cell set \(\ports_T\) is the \(4b_\ell-4\) perimeter cells of its geometric boundary, whether those cells are currently passable or not.
\end{definition}

Because the boundary geometry is fixed, occupancy updates modify only stored metric values, and interface dimensions, matrix offsets, and child-to-parent index mappings remain unchanged.

Write \(T\) for a tile of any kind, whether a base tile of side \(b\), a super-tile assembled from \(g^2\) children, or the root that spans the workspace.
A parent super-tile is denoted \(U\) and one of its children is denoted \(Q\) whenever both must be named together.

The matrix stored at a tile begins as an adjacency.
On a base tile \(T\), \(A_T\) is the weighted adjacency on the vertex set \(V_T\) of all \(b^2\) geometric cells of \(T\),
including currently unknown or occupied slots, so a missing or blocked step is an infinite
entry and a free adjacent step is a finite weight.
On a super-tile \(U\), every path that leaves a child does so through that child's boundary,
and \(A_U\) is the weighted adjacency of the interface graph \(\Gamma_U=(V_U,E_U)\), whose
vertices \(V_U\) are those children's boundary cells placed side by side.
The Kleene star of Section~\ref{sec:kleene} then turns either adjacency into pairwise distances on that same vertex set,
\begin{equation}
  \label{eq:tile_closure}
  C_T=A_T^*,\qquad S_T=C_T\big|_{(\ports_T)\times(\ports_T)},
\end{equation}
so \(A\) is a local adjacency and \(C\) is its Kleene-star closure at every level, while the
vertex set of the matrix is \(V_T\) on a base tile and \(V_U\) on a super-tile.

A base tile therefore carries two cell sets (Fig.~\ref{fig:hierarchy}(\textbf{a})).
The closure \(C_T\) lives on \(V_T\) and answers tile-confined questions: each entry \(C_T(i,j)\) is the length of a shortest path that stays \emph{inside} \(T\), which can exceed the planner geodesic on \(G_\tau\) when a globally shorter route leaves \(T\) and re-enters.
The second set is the boundary \(\ports_T\), the \(4b-4\) perimeter cells.
The parent of \(T\) reads only that perimeter, so the object sent upward is the summary \(S_T\), the restriction of \(C_T\) to the Cartesian product \((\ports_T)\times(\ports_T)\) of those perimeter indices.
The stored tile matrices change with occupancy and would therefore carry the same index \(\tau\) as \(G_\tau\).
We omit that subscript on \(C_T\) and \(S_T\), keeping \(\tau\) on \(G_\tau\) and \(D_\tau\), so the tile symbols stay uncluttered.

A super-tile \(U\) has three cell sets (Fig.~\ref{fig:hierarchy}(\textbf{b})).
The first set is every geometric cell of the square of \(U\), and interior distances on that square are recovered from the two stored sets that follow.
The second set is the interface vertex set
\begin{equation}
  \label{eq:V_U}
  V_U=\bigsqcup_{Q} B_{Q\to U},\qquad B_{Q\to U}:=\varphi_{Q\to U}(\ports_Q),
\end{equation}
formed by the fixed map \(\varphi_{Q\to U}\) that plants each child's boundary \(\ports_Q\) at a contiguous offset \(B_{Q\to U}\) inside \(U\).
In other words, \(V_U\) is the concatenated list of all child boundary cells used as the row
and column indices of \(A_U\) and \(C_U\), and storing the closure \(C_U\) on that list is the
compression that makes the hierarchy possible.
The interface \(\Gamma_U=(V_U,E_U)\) puts a weighted edge between every pair in \(B_{Q\to U}\) whose weight is the matching entry of the child's summary, and seam edges join spatially adjacent perimeter cells of neighboring children when both cells are free.
The closure on that interface is \(C_U=A_U^*\).
The third set is \(\ports_U\subset V_U\), the geometric perimeter of \(U\).
The cells where two children meet, the interior seams, stay in \(V_U\) so that \(C_U\) can
route across children, and they stay out of \(\ports_U\), so the summary \(S_U\) that a further
parent would read lives only on the outer boundary.
Thus \(C_U\) lives on \(V_U\), and the parent-visible summary \(S_U\) lives only on \(\ports_U\subset V_U\).
The geometric cells that fill the square of \(U\) are never stored at \(U\), neither as an adjacency nor as a closure making the hierarchy and compression possible.
Fig.~\ref{fig:hierarchy} takes \(b=5\) and \(g=2\).
A base tile then stores \(C_T\) as a \(25\times 25\) matrix on its 25 cells and exposes \(S_T\) as a
\(16\times 16\) matrix on \(\ports_T\).
The parent concatenates the four 16-cell child boundaries into the 64-index list \(V_U\), stores
\(C_U\) as a \(64\times 64\) matrix on that list, and exposes the 36-cell outer boundary \(\ports_U\).

The root is a super-tile \(U\) with an interface adjacency \(A_U\) on the children's boundaries and a closure \(C_U\) on that interface.
The root stores \(C_U\) on \(V_U\), and a query that is not finalized lower in the tree meets there.
At initialization, each base-tile closure is the Kleene star~\Eqref{eq:kleene} of the currently known occupancy adjacency, after which parent interface graphs are formed bottom-up from child summaries and seam edges and their closures are computed recursively until the root is reached.

\subsection{Min-Plus Updates on Leaves and Ancestors}

An opening \((\Delta V,\Delta E)\) of Section~\ref{sec:problem} is converted into local adjacency decreases, with \(\Delta V=V_{\tau+1}\setminus V_\tau\) the newly free cells and \(\Delta E\) the incident pairs that become finite when those cells are revealed.
Each such pair is written into the decrease set \(\Delta A_T\) of the tile that owns it: an intra-tile edge into its base tile, a seam into the super-tile whose immediate children first separate the endpoints.
We represent \(\Delta A_T\) as the set of those local edge-weight decreases \((a,b,\bar w)\), where \(a\) and \(b\) are the two endpoint cells and \(\bar w\) is the new finite weight.
At a base tile, the decreases are physical grid edges revealed by sensing.
At a super-tile \(U\), the same opening appears as decreased child summaries \(\Delta S_Q\) together with newly enabled seam edges, which together form \(\Delta A_U\).
Whether \(T\) is a base tile or a super-tile, a local adjacency decrease \(\Delta A_T\) first updates that tile's closure \(C_T\).
If this changes any boundary-to-boundary distance, only those changed distances are written into \(\Delta S_T\) and mapped into the parent as \(\Delta A_U\), so that every tile follows \(\Delta A_T\to C_T\to\Delta S_T\to\Delta A_U\).

H-TRIP restores the exactness of the stored closure after those decreases.
The procedure that does this work is \RepairClosure\((C,\Delta A)\).
That procedure takes an already exact closure \(C\) and returns the exact closure of the updated graph.
\(A_T\) is the conceptual local adjacency of the tile.
\(\Delta A_T\) names the direct-edge decreases of that adjacency, and \RepairClosure{} applies those decreases to the stored closure \(C_T\).
\RepairClosure{} first seeds every decreased undirected edge into \(C\) by
\begin{equation*}
  \begin{aligned}
    C(a,b) & \leftarrow\min\bigl\{C(a,b),\bar w\bigr\}, \\
    C(b,a) & \leftarrow\min\bigl\{C(b,a),\bar w\bigr\},
  \end{aligned}
\end{equation*}
and collects those endpoints into an internal set \(\Pi\).
It then performs the min-plus pivot update
\begin{equation*}
  C(i,j)\leftarrow\min\bigl\{C(i,j),\,C(i,\pi)+C(\pi,j)\bigr\}
\end{equation*}
for every \(\pi\in\Pi\), writing the result back into \(C\).
A \emph{dirty tile} is one whose local decrease set \(\Delta A_T\) is nonempty.
Algorithm~\ref{alg:update} repairs every dirty base tile, then every dirty super-tile bottom-up.
All decreases received by the same super-tile from its children, whether those children are base tiles or super-tiles, are first coalesced into one \(\Delta A_U\), after which that dirty parent is repaired once.

\begin{algorithm2e}[t]
  \caption{AbsorbOpening}
  \label{alg:update}
  \footnotesize
  \KwIn{Hierarchical index, opening \((\Delta V,\Delta E)\)}
  \KwOut{Updated exact closures on all affected tiles}
  Add each newly enabled intra-tile edge of \(\Delta E\) to \(\Delta A_T\) of its base tile\;
  Add each seam to \(\Delta A_U\) of the smallest super-tile whose distinct immediate children contain the endpoints\;
  \ForEach{dirty base tile \(T\)}{
  \RepairClosure\((C_T,\Delta A_T)\)\;
  \(\Delta S_T\leftarrow\) entries of \(C_T|_{(\ports_T)\times(\ports_T)}\) that decreased relative to stored \(S_T\)\;
  Write those new values into \(S_T\)\;
  Map every decreased pair \((p,q,w)\in\Delta S_T\) to \((\varphi_{T\to U}(p),\varphi_{T\to U}(q),w)\) and add it to \(\Delta A_U\) for \(U=\mathrm{parent}(T)\)\;
  }
  \ForEach{dirty super-tile \(U\) bottom-up}{
  \RepairClosure\((C_U,\Delta A_U)\)\;
  \If{\(U\) is not the root}{
  \(\Delta S_U\leftarrow\) entries of \(C_U|_{(\ports_U)\times(\ports_U)}\) that decreased relative to stored \(S_U\)\;
  Write those new values into \(S_U\)\;
  Map every decreased pair \((p,q,w)\in\Delta S_U\) to \((\varphi_{U\to\mathrm{parent}(U)}(p),\varphi_{U\to\mathrm{parent}(U)}(q),w)\) and add it to the parent's \(\Delta A\)\;
  }
  }
\end{algorithm2e}
\input{figures/fig-alg-update}

Algorithm~\ref{alg:update} absorbs the opening by that assignment, and the single-cell case is \(|\Delta V|=1\) (Fig.~\ref{fig:alg-update}).
If an exposed summary entry \(S_T(p,q)\) decreases to a finite weight \(w\), the parent
receives the edge decrease \((\varphi_{T\to U}(p),\varphi_{T\to U}(q),w)\) in \(\Delta A_U\).
A seam edge is stored at the first super-tile at which its two endpoint cells belong to different immediate children.
For neighboring base tiles this is their first common parent (Fig.~\ref{fig:hierarchy}(\textbf{b})), so a seam and a child-summary decrease are both adjacency decreases of that parent.
The root is repaired like every other super-tile, but propagation ends there because the root has no parent.

The stored closure \(C\) is the Kleene star of the local adjacency.
That star can be rebuilt from the updated adjacency by unfolding~\Eqref{eq:kleene}, by a Floyd--Warshall pass over every cell, or by a shortest-path search from every cell of the tile.
Each of those reconstructions starts from the adjacency rather than from the already exact \(C\).
Instead, one decreased edge updates the existing closure by the following rank-one identity,
\begin{equation}
  \label{eq:rank1_leaf}
  \begin{aligned}
    C'(i,j) = \min \big\{ & C(i,j),                          \\
                          & C(i,a) + \bar w + C(b,j),        \\
                          & C(i,b) + \bar w + C(a,j) \big\}.
  \end{aligned}
\end{equation}
\addtocounter{equation}{-1}\refstepcounter{equation}\label{eq:conway_rank1}
The first term keeps the old distance, and the other two terms test paths that use the decreased undirected edge once in either direction.
That identity restores one decreased edge, and a batch of decreased edges is restored by the equivalent endpoint-pivot sweep over \(\Pi\) that \RepairClosure{} applies.
A cubic all-pairs pass would treat every cell of a leaf as an intermediate vertex and would cost \(O(b^6)\).
A leaf has \(b^2\) vertices, so that cubic cost is \((b^2)^3\), whereas \(\lvert\Pi_T\rvert\) restricted pivots cost \(\lvert\Pi_T\rvert(b^2)^2\).
The sweep uses only the endpoints in \(\Pi_T\), so the live kernel costs \(O(|\Pi_T|\,b^4)\) on a leaf.
On a \(4\)-connected grid \(|\Pi_T|=O(m_T)\) with \(m_T=\lvert\Delta V\cap V_T\rvert\), so that cost is \(O(m_T b^4)\).
The same restriction at an ancestor replaces a cubic pass of cost \(O(|V_U|^3)\) by \(O(|\Pi_U|\,|V_U|^2)\) on dirty interfaces only.

\noindent\textbf{Blockages.}
\ivparlabel{sec:blockages}
\label{rem:dynamic}
Free-to-obstacle changes can increase distances and therefore require rebuilding the affected leaf and ancestor interfaces, which places them outside the insertion-only update considered here.

\subsection{Batched \(k\times k\) Query}
\label{sec:query_lifting}
A shortest path may leave a leaf and re-enter, so even two viewpoints in the same tile can require ancestor meets.
A queried cell \(p\) first records its distances to the boundary of its base tile, a lift converts
those distances into distances to the boundary of a parent, and two queried cells meet when
their lifted distances are joined through the stored closure of a common ancestor.

Write \(x_{p\to T}\) for the row of distances from \(p\) to \(\ports_T\), with
\begin{equation}
  \label{eq:leaf_attach}
  x_{p\to T}=C_T(p,\ports_T)
\end{equation}
on the leaf that contains \(p\), and write \(s\) and \(t\) only when a pair is formed.
On Fig.~\ref{fig:hierarchy}(\textbf{a}) that row has 16 entries, and a lift through
Fig.~\ref{fig:hierarchy}(\textbf{b}) produces a 36-entry row on \(\ports_U\).
Suppose \(p\) currently lies in child \(Q\) of super-tile \(U\).
If \(Q\) has \(|B_Q|\) boundary cells and \(U\) has \(|B_U|\) exterior boundary cells, then \(x_{p\to Q}\) is a \(1\times|B_Q|\) row, the restriction of \(C_U\) to \((B_{Q\to U})\times(B_U)\) is a \(|B_Q|\times|B_U|\) matrix, and the product \(x_{p\to U}\) is a \(1\times|B_U|\) row.
The corresponding rows of \(C_U\) tell how to continue from the embedded child boundary
\(B_{Q\to U}\) to the outer boundary \(\ports_U\),
\begin{equation}
  \label{eq:source_lift}
  x_{p\to U} = \bigl(x_{p\to Q}\bigr) \otime \Bigl(C_U\big|_{(B_{Q\to U})\times(\ports_U)}\Bigr).
\end{equation}
The entries of \(x_{p\to Q}\) are read in the same order after the fixed mapping \(\varphi_{Q\to U}\),
so they align with the rows indexed by \(B_{Q\to U}\).
Because every stored closure is symmetric, the distances from \(\ports_T\) to \(p\) are the
transpose \(x_{p\to T}^{\top}\).

For an enclosing super-tile \(U\), let \(Q_s\) and \(Q_t\) be the children that contain \(s\) and \(t\).
The meet
\begin{equation}
  \label{eq:ancestor_meet}
  d_U(s,t) = \bigl(x_{s\to Q_s}\bigr) \otime \Bigl(C_U\big|_{(B_{Q_s\to U})\times(B_{Q_t\to U})}\Bigr) \otime \bigl(x_{t\to Q_t}^{\top}\bigr)
\end{equation}
has three factors that respectively represent travel from \(s\) to the boundary of its child, travel across the parent interface, and travel from the target-child boundary to \(t\).
It is the shortest path constrained to that envelope.
At the root the same meet uses the two embedded child boundaries of \(V_U\).
If \(s\) and \(t\) lie in different leaves \(T_s\) and \(T_t\), the first possible meet is their lowest common ancestor.
If they lie in the same leaf, \(C_{T_s}(s,t)\) is also a candidate, and a shorter path may still
leave the leaf and meet at a higher ancestor.
The query therefore takes
\begin{equation}
  \label{eq:global_meet}
  d(s,t) = \begin{cases}
    \min_U d_U(s,t),                                  & T_s \ne T_t, \\[2pt]
    \min \bigl( C_{T_s}(s,t), \min_U d_U(s,t) \bigr), & T_s = T_t,
  \end{cases}
\end{equation}
where the first minimum runs over every super-tile that contains both leaves and the inner minimum
runs over every super-tile that strictly contains \(T_s\).

An isolated pair query may be slower than a single BFS because the hierarchical lift cannot be amortized, and H-TRIP's query advantage therefore comes from amortizing shared lifts over the complete \(k\times k\) matrix.
A single pair costs \(O(H\cdot|B_{\max}|^2)\) with \(|B_{\max}|\) the largest boundary encountered by either query attachment.
At a given level the children that contain query viewpoints share a uniform boundary size \(|B_Q|\), and batched work at that level is \(O(k |B_Q|^2+k^2 |B_Q|)\), of which the first term is the shared lifting work and the second is the pairwise meet work.

\noindent\textbf{Batched matrix.}
\ivparlabel{sec:batched_query}
Algorithm~\ref{alg:query} fills \(D_\tau\in\overline{\mathbb{R}}_{\ge 0}^{k\times k}\) from \(P_\tau=(P_{\tau,1},\dots,P_{\tau,k})\) in one bottom-up sweep.
Cross-child meets share projections.
Every level contributes additional path candidates, and no candidate is assumed final until all relevant ancestors have been considered.
Each meet therefore improves the stored entry by taking the minimum of the incumbent value and the new candidate.
Pairs in different children meet at a parent for the first time, and same-child pairs are reconsidered at that parent because they may leave the child and re-enter it.
At every non-root super-tile the attachments are lifted onto the parent boundary before those meets.
At the root \(U\) every still-relevant pair meets on the interface \(V_U\).

\begin{algorithm2e}[t]
  \caption{Batched \(k\times k\) query}
  \label{alg:query}
  \footnotesize
  \KwIn{Hierarchy, viewpoint list \(P_\tau\) of length \(k\)}
  \KwOut{Matrix \(D_\tau\in\overline{\mathbb{R}}_{\ge 0}^{k\times k}\)}
  Initialize \(D_\tau\leftarrow\infty\); set \((D_\tau)_{ii}\leftarrow 0\) for every passable \(P_{\tau,i}\)\;
  Bucket each \(P_{\tau,i}\) into its base tile \(T_i\) and read \(x_{(P_{\tau,i}\to T_i)}\) from \(C_{T_i}\)\;
  For each pair with \(T_i=T_j\), set \((D_\tau)_{ij}\leftarrow C_{T_i}(P_{\tau,i},P_{\tau,j})\) and likewise \((D_\tau)_{ji}\)\;
  \ForEach{non-root super-tile \(U\) bottom-up}{
  Lift every viewpoint attachment from its child to the parent boundary\;
  Evaluate every relevant pair through the parent interface\;
  Set \((D_\tau)_{ij}\leftarrow\min\{(D_\tau)_{ij},d_U(P_{\tau,i},P_{\tau,j})\}\) and likewise \((D_\tau)_{ji}\)\;
  }
  At the root \(U\), evaluate every still-relevant pair through \(V_U\)\;
  Set \((D_\tau)_{ij}\leftarrow\min\{(D_\tau)_{ij},d_U(P_{\tau,i},P_{\tau,j})\}\) and likewise \((D_\tau)_{ji}\)\;
\end{algorithm2e}

\subsection{Exactness}

\begin{lemma}[Rank-one update]
  \label{lem:confluence}
  Let \(G=(V,E,\omega)\) be undirected with nonnegative weights, and let \(C=A^*\) be its metric closure.
  Inserting an undirected edge \(e=\{u,v\}\) of weight \(\bar w\ge 0\), or decreasing an existing edge to \(\bar w\), produces the new closure~\Eqref{eq:rank1_leaf} for every pair of vertices.
  Sequential application of the rank-one update to every inserted or decreased edge yields the metric closure of the final graph, independent of insertion order.
  Equivalently, after seeding all decreased edges, a min-plus pivot sweep over their endpoint set \(\Pi\) yields the same updated closure.
\end{lemma}
\begin{proof}
  An optimal simple path either avoids the decreased edge or uses it once, in which case the old closure already contains the prefix and suffix of~\Eqref{eq:rank1_leaf}.
  Every new shortest path decomposes into old shortest subpaths joined at vertices in \(\Pi\), so the endpoint-pivot sweep yields the same closure.
\end{proof}

\begin{theorem}[Monotone update exactness]
  \label{thm:monotone_exactness}
  After Algorithm~\ref{alg:update} absorbs an insertion-only opening, every stored tile closure remains exact.
\end{theorem}
\begin{proof}
  Every dirty tile absorbs all decreases in its local \(\Delta A_T\) by Lemma~\ref{lem:confluence}, so that tile's stored closure remains exact.
  If the tile is not the root and the exterior summary \(S_T\) strictly decreases, those entries are written into the parent as \(\Delta A_U\) together with new seams, while internal entries that do not change the exterior summary stay local.
  Induction along each touched ancestor chain, stopping at the root or at an empty \(\Delta S_U\), extends the claim to the tree.
\end{proof}

\begin{theorem}[Hierarchical query exactness]
  \label{thm:hierarchical_preservation}
  Given vertices \(s,t\in V_\tau\), the hierarchical query of Algorithm~\ref{alg:query} returns the exact grid geodesic \(d_{G_\tau}(s,t)\).
\end{theorem}
\begin{proof}
  The attachment of a queried cell to the boundary of its leaf and of every non-root ancestor equals the geodesic on the occupancy subgraph induced by that tile, because every exit is a perimeter cell and every child-summary edge of the interface adjacency \(A_T\) realizes a child-internal path.
  Every true grid path compresses to an equal-length walk on the interface of the smallest ancestor region that contains the complete path, and every interface walk expands through stored child summaries to a realizable grid path, so the minimum of~\Eqref{eq:global_meet} is the geodesic and Algorithm~\ref{alg:query} records \(d_{G_\tau}(s,t)\) for every queried pair.
\end{proof}
Consequently, any surviving-pair shortcut introduced by an opening is reflected in the next query after Algorithm~\ref{alg:update} propagates the corresponding decreases.

\noindent\textbf{Memory.}
\ivparlabel{sec:memory_complexity}
A base tile stores \(b^4\) closure entries and a boundary summary of \((4b-4)^2\) entries.
A level-\(\ell\) tile \(T\) has side \(b_\ell=bg^{\ell}\) and \(|B_T|=4b_\ell-4\), so the parent interface has \(|V_U|=g^2|B_Q|\) when each child \(Q\) is a level-\(\ell\) tile.
The corresponding super-tile closure therefore stores \(|V_U|^2\) entries.

\section{Experimental Evaluation}
\label{sec:experiments}

This section evaluates H-TRIP on two-dimensional occupancy maps from the MovingAI benchmark~\cite{sturtevant2012benchmarks}.
The evaluation covers numerical exactness, batched query performance, incremental update cost, and multi-robot exploration, which together characterize the regime in which the hierarchy is beneficial and the cost of maintaining it as free space is revealed.

\noindent\textit{A.~Setup and Exactness:}
\label{sec:eval_setup}
The evaluation uses two settings.
To assess batched query performance, every ground-truth passable cell of each map is marked observed free, so those queries include all traversable cells.
\label{sec:system}
The second setting places sixteen unicycle robots with onboard two-dimensional lidars, which explore occupancy grids with \(5\,\mathrm{cm}\) cells in the frontier-based explorer of Fig.~\ref{fig:explorer}.
At each planning cycle the occupancy graph \(G_\tau\) consists of the cells the fleet has explored, frontiers are extracted from that occupancy, and robot poses and frontier representatives snap onto free cells of \(G_\tau\).
Each query is exact.
Indexing uses a double buffer so that an update does not block queries.
Table~\ref{tab:maps} names the maps used in both settings.

\begin{table}[t]
    \caption{Evaluation maps from the MovingAI open-source set~\cite{sturtevant2012benchmarks}}
    \label{tab:maps}
    \centering
\begingroup
\setlength{\fboxsep}{0pt}%
\setlength{\fboxrule}{0.25pt}%
\setlength{\tabcolsep}{2.4pt}%
\newcommand{\mapimg}[1]{%
    \fcolorbox{black!40}{white}{\includegraphics[width=0.148\columnwidth,interpolate=false]{figures/thumbs/#1.png}}}%
\resizebox{\columnwidth}{!}{%
\begin{tabular}{@{}cccccc@{}}
\mapimg{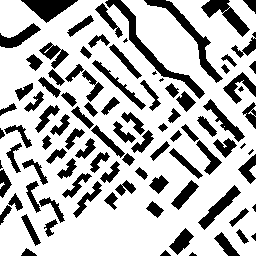} & \mapimg{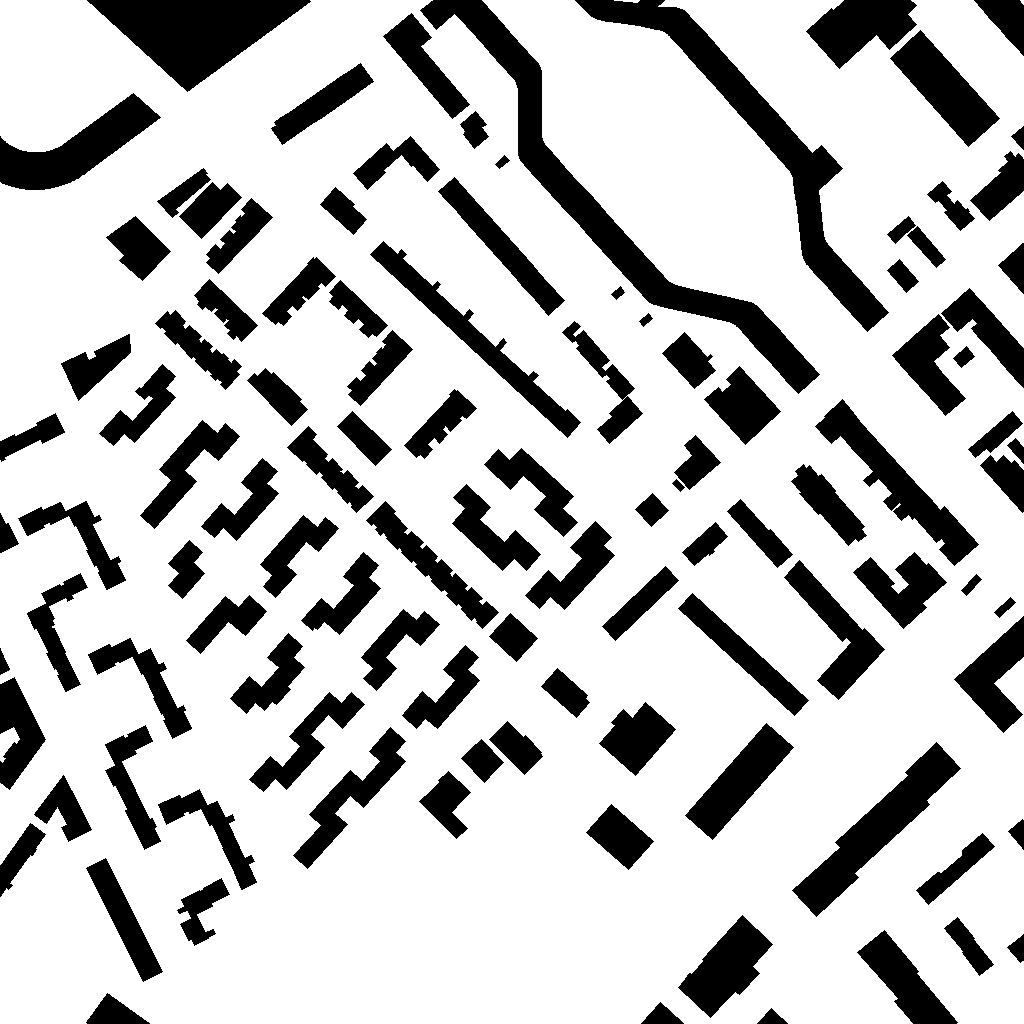} & \mapimg{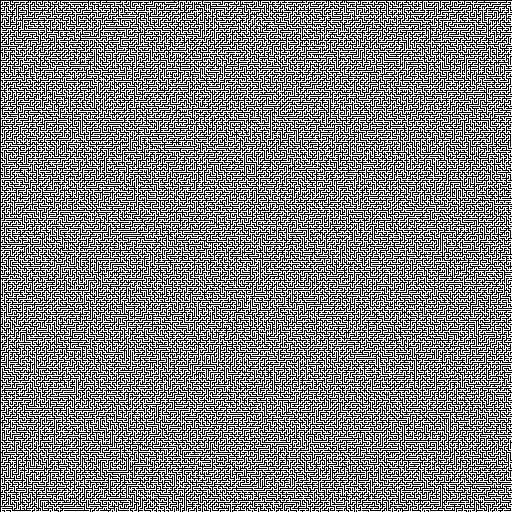} & \mapimg{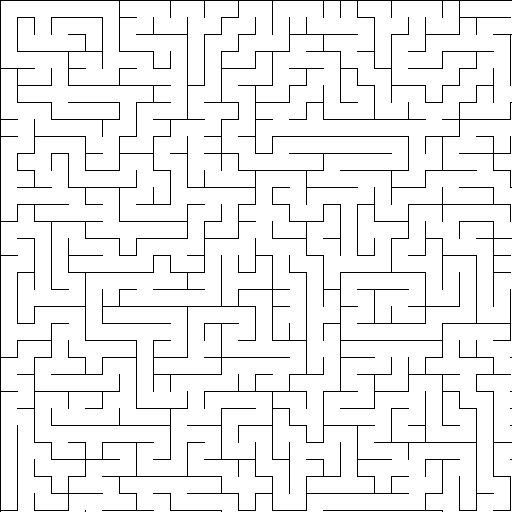} & \mapimg{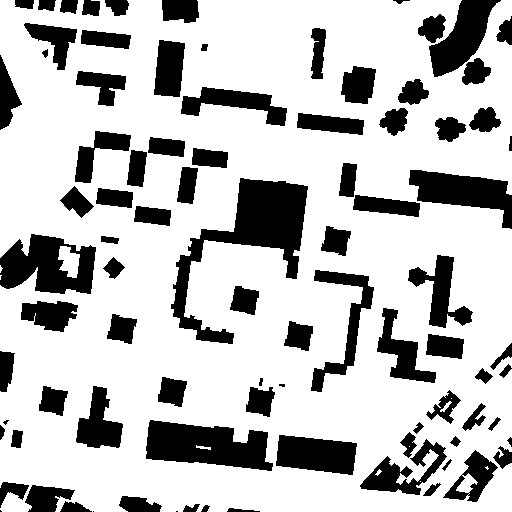} & \mapimg{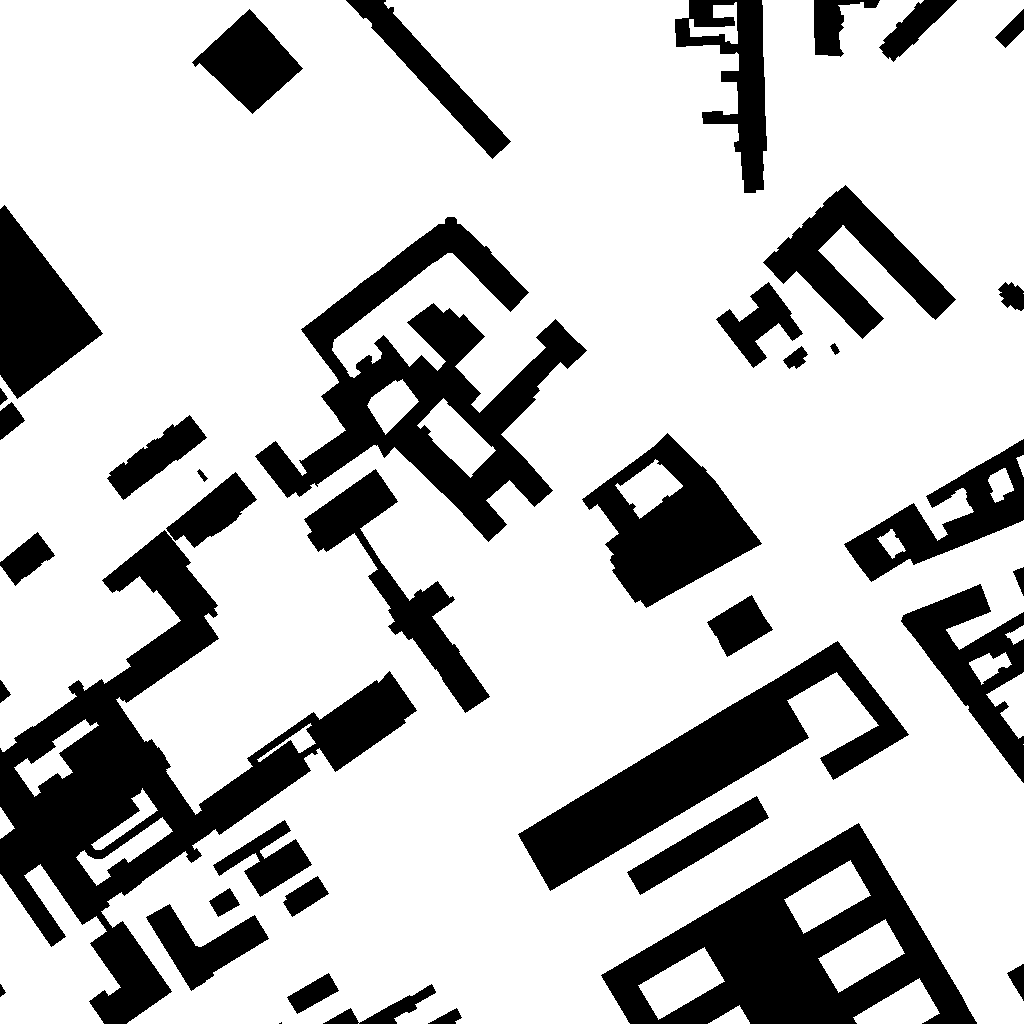} \\[-1.5pt]
{\scriptsize\Map{1}} &
{\scriptsize\Map{2}} &
{\scriptsize\Map{3}} &
{\scriptsize\Map{4}} &
{\scriptsize\Map{5}} &
{\scriptsize\Map{6}}
\end{tabular}}%
\endgroup
\vspace{3.5pt}

    \scriptsize
    \setlength{\tabcolsep}{3.2pt}
    \resizebox{\columnwidth}{!}{%
        \begin{tabular}{@{}llcr@{\gvr}llcr@{}}
            \toprule
            Map     & Name                     & \(b,g\)  & \(\lvert V\rvert\) &
            Map     & Name                     & \(b,g\)  & \(\lvert V\rvert\)   \\
            \midrule
            \Map{1} & \texttt{Boston\_0\_256}  & \(16,2\) & \(47{,}768\)       &
            \Map{4} & \texttt{maze512-16-0}    & \(16,2\) & \(246{,}016\)        \\
            \Map{2} & \texttt{Boston\_0\_1024} & \(32,2\) & \(796{,}896\)      &
            \Map{5} & \texttt{Paris\_0\_512}   & \(8,2\)  & \(196{,}567\)        \\
            \Map{3} & \texttt{maze512-1-0}     & \(16,2\) & \(131{,}071\)      &
            \Map{6} & \texttt{Berlin\_0\_1024} & \(8,2\)  & \(794{,}748\)        \\
            \bottomrule
        \end{tabular}}
\end{table}

In that first setting, query points are selected homogeneously across each map.
The timed \(k=25\) batch uses twenty-five of those points, and each reported query time is the median of ten timed repeats after three cache warm-ups.
\label{sec:avx2}
Min-plus kernels use AVX2 on an Intel Core i7-8650U, and \(k\times\)BFS is a scalar, single-threaded search on the same graph.
Pairwise \(\mathrm{A}^*\) at \(k=25\) is \(7.3\times\)--\(9.9\times\) slower than that \(k\times\)BFS fill on \Map{1}, \Map{2}, and \Map{3}, and the query tables report \(k\times\)BFS as the sequential-search baseline.
Distances in the index are \(\texttt{uint16}\) with sentinel \(\texttt{INF}=60000\), which is sufficient for the maps in this evaluation.
Tables~\ref{tab:prelim_k25} and~\ref{tab:query_vs_k} time Algorithm~\ref{alg:query} on those fully revealed maps, and Table~\ref{tab:closedloop} times the same \(k\times k\) fill on the occupancy graph \(G_\tau\) as the robots explore.
\label{sec:eval_exactness}
Across \(178{,}708\) logged matrix entries compared against BFS or Dijkstra, H-TRIP produced zero mismatches.

\noindent\textit{B.~Batched Query Performance:}
\label{sec:prelim}
\begin{table}[t]
    \caption{Median query time for \(25\) viewpoints on a fully revealed map.
        Index is the size of the H-TRIP index.
        Break-even is the number of batches that recover the full-build cost against \(k\times\)BFS.}
    \label{tab:prelim_k25}
    \centering
    \scriptsize
    \setlength{\tabcolsep}{3.0pt}
    \begin{tabular}{@{}l@{\gvr}r@{\gvr}r@{\gvr}r@{\gvr}r@{\gvr}r@{\gvr}r@{}}
        \toprule
        Map     & Index                 & Full build           & \(k\times\)BFS         & H-TRIP                & vs BFS         & Break-even \\
        \midrule
        \Map{1} & \(87\,\mathrm{MB}\)   & \(0.71\,\mathrm{s}\) & \(32.7\,\mathrm{ms}\)  & \(3.3\,\mathrm{ms}\)  & \(9.9\times\)  & \(24\)     \\
        \Map{2} & \(3175\,\mathrm{MB}\) & \(79.9\,\mathrm{s}\) & \(554.2\,\mathrm{ms}\) & \(49.8\,\mathrm{ms}\) & \(11.1\times\) & \(158\)    \\
        \Map{3} & \(404\,\mathrm{MB}\)  & \(1.07\,\mathrm{s}\) & \(149.4\,\mathrm{ms}\) & \(1.81\,\mathrm{ms}\) & \(82.6\times\) & \(7\)      \\
        \Map{4} & \(404\,\mathrm{MB}\)  & \(3.40\,\mathrm{s}\) & \(207\,\mathrm{ms}\)   & \(18.0\,\mathrm{ms}\) & \(11.5\times\) & \(18\)     \\
        \Map{5} & \(351\,\mathrm{MB}\)  & \(6.33\,\mathrm{s}\) & \(166\,\mathrm{ms}\)   & \(17.0\,\mathrm{ms}\) & \(9.77\times\) & \(42\)     \\
        \Map{6} & \(1629\,\mathrm{MB}\) & \(63.0\,\mathrm{s}\) & \(744\,\mathrm{ms}\)   & \(102\,\mathrm{ms}\)  & \(7.28\times\) & \(98\)     \\
        \bottomrule
    \end{tabular}
\end{table}
\begin{table*}[t]
    \centering
    \scriptsize
    \newcommand{\EvalTabRowScale}{0.86}
    \begin{minipage}[t]{0.325\textwidth}
        \centering
        \caption{Speedup of H-TRIP versus \(k\times\)BFS on fully revealed maps for various \(k\) values.
            The large ratios on \Map{3} come from one-cell corridors.}
        \label{tab:query_vs_k}
        \setlength{\tabcolsep}{2.6pt}
        \scalebox{\EvalTabRowScale}{%
            \begin{tabular}{@{}l@{\gvr}c@{\gvr}c@{\gvr}c@{\gvr}c@{\gvr}c@{\gvr}c@{}}
                \toprule
                Map     & \(k{=}2\)      & \(k{=}4\)      & \(k{=}16\)     & \(k{=}32\)     & \(k{=}64\)     & \(k{=}256\)    \\
                \midrule
                \Map{1} & \(0.67\times\) & \(1.70\times\) & \(8.87\times\) & \(8.38\times\) & \(7.04\times\) & \(1.79\times\) \\
                \Map{2} & \(0.51\times\) & \(2.41\times\) & \(13.7\times\) & \(8.97\times\) & \(7.41\times\) & \(2.00\times\) \\
                \Map{3} & \(120\times\)  & \(89.2\times\) & \(235\times\)  & \(67.2\times\) & \(54.6\times\) & \(11.4\times\) \\
                \Map{4} & \(1.50\times\) & \(3.98\times\) & \(24.6\times\) & \(8.26\times\) & \(13.0\times\) & \(2.92\times\) \\
                \Map{5} & \(0.12\times\) & \(0.70\times\) & \(6.71\times\) & \(8.48\times\) & \(7.52\times\) & \(2.56\times\) \\
                \Map{6} & \(0.10\times\) & \(0.66\times\) & \(5.94\times\) & \(7.24\times\) & \(2.31\times\) & \(1.82\times\) \\
                \bottomrule
            \end{tabular}}
    \end{minipage}\hfill
    \begin{minipage}[t]{0.325\textwidth}
        \centering
        \caption{Index update after occupancy openings.
            Upd.\ is Algorithm~\ref{alg:update}, Upd.+query adds Algorithm~\ref{alg:query}, and \(k\times\)BFS \(D_\tau\) is a refill.}
        \label{tab:update}
        \setlength{\tabcolsep}{2.4pt}
        \scalebox{\EvalTabRowScale}{%
            \begin{tabular}{@{}l@{\gvr}l@{\gvr}r@{\gvr}r@{\gvr}r@{\gvr}r@{}}
                \toprule
                Map     & Pattern    & \(m\)  & Upd.                   & Upd.+query            & \(k\times\)BFS \(D_\tau\) \\
                \midrule
                \Map{1} & same-leaf  & \(1\)  & \(0.042\,\mathrm{ms}\) & \(3.34\,\mathrm{ms}\) & \(34.6\,\mathrm{ms}\)     \\
                        & same-leaf  & \(16\) & \(74.8\,\mathrm{ms}\)  & \(78.1\,\mathrm{ms}\) & \(33.3\,\mathrm{ms}\)     \\
                        & multi-leaf & \(64\) & \(242\,\mathrm{ms}\)   & \(245\,\mathrm{ms}\)  & \(36.2\,\mathrm{ms}\)     \\
                \Map{3} & multi-leaf & \(64\) & \(154\,\mathrm{ms}\)   & \(156\,\mathrm{ms}\)  & \(174\,\mathrm{ms}\)      \\
                \Map{4} & same-leaf  & \(1\)  & \(0.185\,\mathrm{ms}\) & \(18.2\,\mathrm{ms}\) & \(252\,\mathrm{ms}\)      \\
                        & same-leaf  & \(16\) & \(3.44\,\mathrm{ms}\)  & \(21.4\,\mathrm{ms}\) & \(258\,\mathrm{ms}\)      \\
                \bottomrule
            \end{tabular}}
    \end{minipage}\hfill
    \begin{minipage}[t]{0.325\textwidth}
        \centering
        \caption{Query times during exploration (\(N{=}16\), \(K{=}50\), \( \le 500\) sim. steps) on map revealed so far.
            LiDAR is range, span, and ray count.}
        \label{tab:closedloop}
        \setlength{\tabcolsep}{2.2pt}
        \scalebox{\EvalTabRowScale}{%
            \begin{tabular}{@{}l@{\gvr}c@{\gvr}c@{\gvr}c@{\gvr}r@{\gvr}r@{\gvr}r@{}}
                \toprule
                Map     & LiDAR                          & \(b\)  & Cov.       & H-TRIP                & \(k\times\)BFS        & vs BFS         \\
                \midrule
                \Map{1} & \(6\,\mathrm{m}/180^\circ/45\) & \(16\) & \(93.3\%\) & \(14.2\,\mathrm{ms}\) & \(122\,\mathrm{ms}\)  & \(8.59\times\) \\
                \Map{4} & \(4\,\mathrm{m}/90^\circ/30\)  & \(16\) & \(27.1\%\) & \(6.08\,\mathrm{ms}\) & \(16.7\,\mathrm{ms}\) & \(2.75\times\) \\
                \Map{4} & \(8\,\mathrm{m}/180^\circ/60\) & \(16\) & \(39.1\%\) & \(6.37\,\mathrm{ms}\) & \(23.0\,\mathrm{ms}\) & \(3.61\times\) \\
                \Map{5} & \(5\,\mathrm{m}/90^\circ/30\)  & \(8\)  & \(68.7\%\) & \(56.9\,\mathrm{ms}\) & \(239\,\mathrm{ms}\)  & \(4.20\times\) \\
                \Map{5} & \(5\,\mathrm{m}/90^\circ/30\)  & \(32\) & \(69.5\%\) & \(63.2\,\mathrm{ms}\) & \(212\,\mathrm{ms}\)  & \(3.35\times\) \\
                \Map{6} & \(5\,\mathrm{m}/90^\circ/30\)  & \(8\)  & \(50.2\%\) & \(216\,\mathrm{ms}\)  & \(478\,\mathrm{ms}\)  & \(2.21\times\) \\
                \bottomrule
            \end{tabular}}
    \end{minipage}
\end{table*}
\begin{figure*}[t]
    \centering
    \setlength{\abovecaptionskip}{0pt}
    \begin{minipage}[t]{0.495\textwidth}
        \centering
        \includegraphics[width=\linewidth]{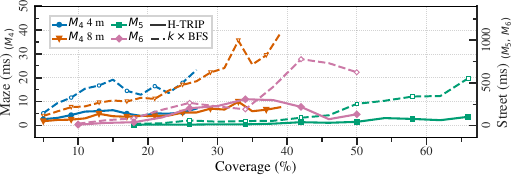}
        \captionof{figure}{Free-cell coverage versus paired query time during exploration.
            Across the traces, \(k\times\)BFS query time generally grows faster with coverage than H-TRIP.}
        \label{fig:coverage_query}
    \end{minipage}\hfill
    \begin{minipage}[t]{0.495\textwidth}
        \centering
        \includegraphics[width=\linewidth]{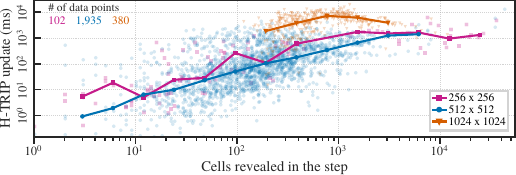}
        \captionof{figure}{H-TRIP update time versus cells revealed in the same exploration step, grouped by native map side.
            Markers are changed-step telemetry samples, and curves are log-spaced bin medians.}
        \label{fig:opening_update}
    \end{minipage}
\end{figure*}

H-TRIP is \(7.28\times\)--\(11.5\times\) faster than \(k\times\)BFS at \(k=25\) on \Map{1}, \Map{2}, \Map{4}, \Map{5}, and \Map{6}.
\Map{3} is a one-cell-wide corridor maze whose query points lie thousands of cells apart, so each BFS walks a long unique path and the paired median ratio reaches \(82.6\times\).
The full-build cost is amortized after twenty-four batches on \Map{1} and seven on \Map{3}.

On \Map{3} the one-cell corridors keep each BFS on a long unique path, so the ratios stay large across the sweep (Table~\ref{tab:query_vs_k}).
On the street maps H-TRIP overtakes \(k\times\)BFS once the batch shares hierarchical lifts, and the advantage then declines at very large \(k\) because the \(k\times k\) lift has quadratic meet cost.
The benefit is therefore workload dependent, because H-TRIP reuses hierarchical lifts when viewpoints are spatially distributed.

\noindent\textit{C.~Incremental Update Performance:}
\label{sec:eval_update}
Local openings are cheap when propagation stops near the leaf, and they become expensive when decreases dirty several ancestors, which is map dependent (Table~\ref{tab:update}).
A broad multi-leaf batch can exceed a BFS refill of \(D_\tau\), so the exploration experiment tests whether the cheap local regime occurs often enough as the fleet opens cells.

\noindent\textit{D.~Multi-Robot Exploration:}
\label{sec:eval_closed_loop}
The runs in Table~\ref{tab:closedloop} place sixteen robots on the occupancy graph \(G_\tau\) that the fleet has explored so far, and each paired sample fills \(D_\tau\) at those viewpoints.
A run stops earlier when free-cell coverage reaches \(95\%\).
When coverage is still partial, \(k\times\)BFS searches a smaller graph than the fully revealed maps of Table~\ref{tab:prelim_k25}, so the exploration speedup is smaller than the full-map figure.
On \Map{1} coverage is nearly complete, so the ratio stays near that full-map speedup.
On \Map{4} a wider lidar raises coverage, so \(k\times\)BFS takes longer, and H-TRIP stays cheap because most of the maze is still unknown (Fig.~\ref{fig:coverage_query}).
On \Map{5} a larger leaf reduces the paired speedup, and on \Map{6} occupancy is still partial while H-TRIP remains faster than \(k\times\)BFS.
Fig.~\ref{fig:opening_update} plots Algorithm~\ref{alg:update} times against cells revealed in the same step, grouped by map side, and each curve pools in-loop updates at that resolution.

\section{Conclusion}
\label{sec:conclusion}

This paper introduced H-TRIP, an exact hierarchical distance index for multi-robot exploration on monotonically growing occupancy grids.
By combining fixed boundary ports, min-plus closure, and localized rank-one updates, H-TRIP maintains exact geodesic distances while supporting repeated many-to-many evaluation as the map expands.
During multi-robot exploration, batched queries were \(2.21\times\)--\(8.59\times\) faster than co-scheduled \(k\times\)BFS while preserving exact agreement with reference searches.
Update cost remains workload dependent, however, and openings that propagate through several hierarchy levels can exceed the cost of a BFS refill.
The current formulation is limited to insertion-only changes in two-dimensional occupancy grids; future work will consider non-monotonic updates, lower-memory representations, and physical multi-robot validation.

\bibliographystyle{IEEEtran}
\bibliography{refs}

\end{document}